%% file: main.tex
\documentclass[11pt,a4paper]{article}
\usepackage[hyperref]{acl2021}
\usepackage{times}
\usepackage{latexsym}

\usepackage{microtype}

\aclfinalcopy 

\usepackage{soul}
\usepackage{url}
\usepackage{graphicx}
\usepackage{amsmath}
\usepackage{amsthm}
\usepackage{xspace}
\usepackage{pifont}
\usepackage{caption}
\usepackage{subcaption}
\usepackage{arydshln}
\usepackage[ruled,noline,linesnumbered]{algorithm2e}
\usepackage{multirow}
\usepackage{xcolor}

\newcommand{\task}{\texttt{proScript}\xspace}
\newcommand{\data}{\texttt{proScript}\xspace}
\newcommand{\modeledge}{\texttt{proScript}$_\text{edge-pred}$\xspace}
\newcommand{\modelgen}{\texttt{proScript}$_\text{gen}$\xspace}
\newcommand{\modelgentransfer}{\texttt{proScript}$_\text{gen-transfer}$\xspace}
\newcommand{\modelgenpipe}{\texttt{proScript}$_\text{gen-pipe}$\xspace}

\definecolor{emerald}{rgb}{0.31, 0.78, 0.47}
\definecolor{piggypink}{rgb}{0.99, 0.87, 0.9}
\definecolor{teagreen}{rgb}{0.82, 0.94, 0.75}

\newcommand{\eat}[1]{}

\usepackage{quoting}

\newenvironment{ite}{                     
     \parskip 0cm \begin{itemize} \parskip 0cm \parsep 0cm \itemsep 0cm \topsep 0cm}{
        \end{itemize}} 

\title{\data: Partially Ordered Scripts Generation \\via Pre-trained Language Models}

\newcommand{\allenai}{$^{\dagger}$}
\newcommand{\uw}{$^{\ddagger}$}
\author{
  Keisuke Sakaguchi\allenai, 
  \quad Chandra Bhagavatula\allenai,
  \quad Ronan Le Bras\allenai, \\
\textbf{Niket Tandon\allenai,
  \quad Peter Clark\allenai,
  \quad Yejin Choi\allenai\uw}
\\
\allenai Allen Institute for AI, \uw University of Washington\\
{\texttt \{keisukes, chandrab, rolanlb, nikett, peterc, yejinc\}@allenai.org}\\
}

\date{}

\begin{document}
\maketitle

\input src/abstract

\input src/introduction

\input src/related

\input src/definition

\input src/datasets

\input src/task-edge

\input src/task-gen

\input src/conclusions

%

\bibliographystyle{acl_natbib}
\bibliography{acl2021}

\appendix
\section{Appendices}
\label{sec:appendix}
\input src/appendices

\end{document}

%% file: src/abstract.tex
\begin{abstract}
{\it Scripts} - standardized event sequences describing typical everyday activities -
have been shown to help understand narratives by providing expectations, 
resolving ambiguity, and filling in unstated information. However, to date
they have proved hard to author or extract from text. In this work,
we demonstrate for the first time that pre-trained neural language models (LMs)
can be be finetuned to {\it generate} high-quality scripts, at varying
levels of granularity, for a wide range of everyday scenarios (e.g., bake a cake). To do this, we 
collected a large (6.4k), crowdsourced partially ordered scripts (named \data), 
which is substantially larger than prior datasets, 
and developed models that generate scripts with combining language generation and structure prediction.
We define two complementary tasks: (i) edge prediction: given a scenario and unordered events,
organize the events into a valid (possibly partial-order) script, and
(ii) script generation: given only a scenario, generate
events and organize them into a (possibly partial-order)
script. Our experiments show that our models perform
well (e.g., F1=75.7 on task (i)),
illustrating a new approach to overcoming previous barriers
to script collection. We also show that there is still significant
room for improvement toward human level performance. Together,
our tasks, dataset, and models offer a new research direction for
learning script knowledge. 

\end{abstract}

%% file: src/introduction.tex
\section{Introduction}
\label{sec:introduction}

\begin{figure}[t]
\begin{center}
\includegraphics[width=0.8\columnwidth]{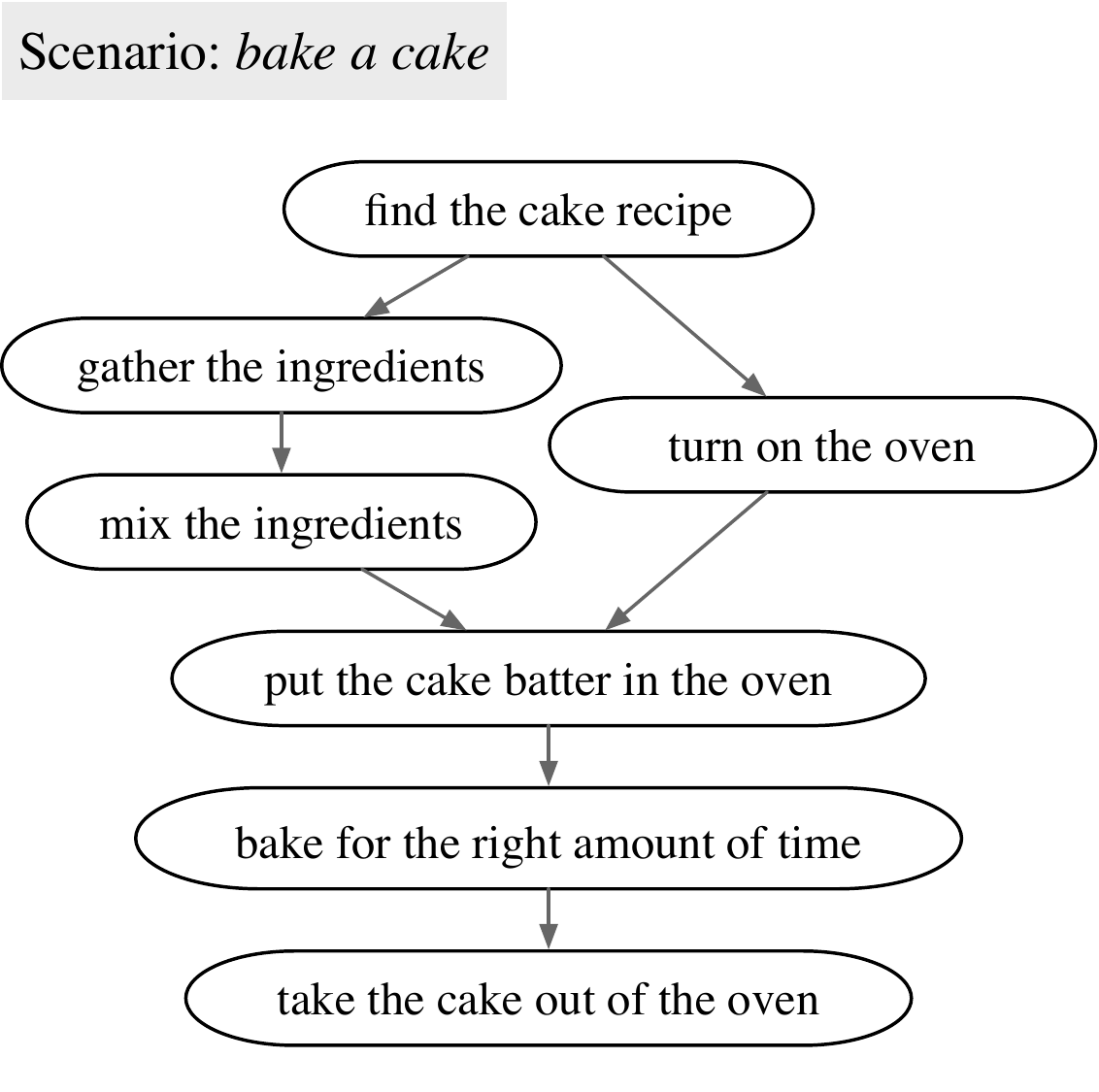}
\caption{We collected 6.4k of partially ordered scripts (\data) and developed models that take a scenario (e.g., bake a cake) as the input and \textit{generate} a (possibly partial-order) script.}
\label{fig:fig1}
\end{center}
\end{figure}

\textit{Scripts}, originally introduced by \newcite{Schank1975ScriptsPA}, represent structured commonsense knowledge about prototypical events in everyday situations/scenarios such as \textit{bake a cake} and \textit{fuel a car} (Figure~\ref{fig:fig1}).
However, while scripts have been shown to help understand narratives by providing expectations,
resolving ambiguity, and filling in unstated information~\cite[inter alia]{chambers-jurafsky-2008-unsupervised,modi-etal-2017-modeling},
they have proved hard to author or extract from text, with only small script databases available~\cite{regneri-etal-2010-learning,chambers-2017-behind,Ostermann_2020}.

In this work, we show for the first time that pre-trained neural language models (LMs) can be adapted to
{\it generate} high-quality scripts, including appropriately partial ordering
events where a specific temporal ordering is required only when it is necessary. 
LMs have previously been shown to successfully generate stories~\cite{rashkin-etal-2020-plotmachines},
summaries~\cite{lewis-etal-2020-bart}, and commonsense facts~\cite{Bosselut2019COMETCT,hwang2020comet}. 
Here we investigate their application to script \textit{generation}.
First, we collect large amount (6.4k) of partially ordered script from crowdsourcing with a similar but simplified collection method~\cite{Ciosici2021MachineAssistedSC}.
We call the dataset as \data (PaRtial Order SCRIPt for generaTion), and this is substantially larger than prior (crowdsourced) dataset such as DeScript~\cite{regneri-etal-2010-learning} that has 40 scripts.
Since the granularity of scripts (and the events) are inherently vague and subjective~\cite{modi-etal-2016-inscript}, we collected wider variety of micro and macroscopic scripts than previous datasets. 
Additionally, temporal duration of each event is also annotated (e.g., \textit{take the cake out of the oven} typically takes one minute in the \textit{bake a cake} script), which will potentially link script knowledge with temporal reasoning in future work.

Second, with the collected data, we introduce two complementary tasks: \textbf{script edge prediction} and \textbf{entire script generation}.
In the edge prediction task, given a scenario and unordered intermediate events, models must organize the events as a valid partial-order script. 
On the other hand, the script generation task is to generate intermediate events and the partial-order of those events for a given scenario. 
This task requires both natural language generation (for nodes) and graph structure prediction (for edges). 

Finally, based on our proposed dataset, we develop models for both edge prediction and entire script generation tasks. 
As \newcite{chambers-2017-behind} has revealed that models trained and evaluated on missing events prediction (i.e., \textit{narrative cloze}) are insufficient to assess script knowledge, our evaluation scheme evaluate the entire script. 
We compare the models against baselines,
and show that our models outperform the baselines for both the edge prediction and the script generation tasks. 
Nonetheless, there is a significant room for improvement toward human-level performance 
-- e.g., for edge prediction, the best model achieves 75.71 of F1 score while human achieves 89.28, and for script generation, the best model obtains a graph edit distance of 4.97 (i.e., number of human edits), while human-created scripts achieve 2.98 on average. 

\noindent
Our contributions are thus:
\begin{ite}
\item A new dataset (\data) of crowdsourced scripts that is substantially larger than prior (manually crafted) datasets
\item Two complementary task definitions against \data
\item Two new models for these task, providing the first demonstration that generative models can
      be successfully applied, although it is still significantly below human levels
\end{ite}

%% file: src/related.tex
\section{Related Work}
\label{sec:related}
\paragraph{Script as narrative chain} 
\newcite{mooney1985learning} and \newcite[inter alia]{chambers-jurafsky-2008-unsupervised} have investigated automatically inducing scripts from (unstructured) corpus. 
In particular, \newcite{chambers-jurafsky-2008-unsupervised} introduced scripts as \textit{narrative chain}, where verbs with the participants information (e.g., (\textit{claimed, subj}), and (\textit{accused, obj}) ) named \textit{narrative events} are partially ordered according to causal and temporal relations.
They also introduced \textit{narrative cloze} task, where a model is expected to predict one removed narrative event, given all the other narrative events, while our proposed task requires to \textit{generate} scripts as a partial-order graph for a given scenario.
The ``script as narrative chain'' approach has been actively studied~\cite{jans-etal-2012-skip,modi-titov-2014-inducing,pichotta-mooney-2014-statistical,rudinger-etal-2015-script,granroth2016happens,weber-etal-2018-hierarchical,belyy-van-durme-2020-script}, but it has its drawbacks.
First, the source corpora is mainly from a news domain rather than everyday scenarios, and a number of verbs in news texts are not script-relevant events such as reporting verbs ~\cite{mostafazadeh-etal-2016-corpus,chambers-2017-behind}.
Second, events are highly abstracted as tuples of verb and the dependency (\textit{subj} or \textit{obj})~\cite{Ostermann_2020}.
Third, the evaluation scheme for the narrative cloze task is insufficient to evaluate script knowledge ~\cite{chambers-2017-behind}.

\paragraph{Script as paraphrase sets}
\textit{Script as paraphrase sets}~\cite{regneri-etal-2010-learning,modi-etal-2016-inscript,wanzare-etal-2016-crowdsourced} is more recent approach to gather script knowledge, where crowd workers are asked to write down a sequence of events for a given everyday scenario (e.g., bake a cake) and the collected sequences (called event sequence description) are aligned with paraphrased events being clustered.
The collected (partially ordered) scripts cover wide variety of everyday situations compared to narrative chains (news domain), but one shortcoming of this approach is the scalability; it is not easy to scale because of the cost for manual data collection~\cite{chambers-2017-behind,Ostermann_2020}. 
In fact, \newcite{modi-etal-2016-inscript} crowdsourced 1000 stories that cover only 10 scripts, and similarly \newcite{regneri-etal-2010-learning} end up with collecting 40 scripts. 
The limited amount of data hinders learning script knowledge by models.
Furthermore, they provide no evaluation metric on the dataset for assessing model's script knowledge.

\paragraph{Story generation}
Neural models have been demonstrated to successfully generate stories~\cite{kiddon-etal-2016-globally,peng-etal-2018-towards,zhai-etal-2019-hybrid,rashkin-etal-2020-plotmachines}. 
Our work is related to story generation in terms of generating higher-level agenda (or plot) of a story. 
However, a main difference between stories and scripts is that stories often require surprising and incidental \textit{sequence} of events as well as description about character's mental states and landscape depiction that make the story attractive for readers, whereas our script generation expects generating essential \textit{core events}~\cite{chambers-2017-behind} in \textit{partial order}.

%% file: src/definition.tex
\section{Definitions}
\label{sec:definition}
\paragraph{\task}
We define \task as a directed acyclic graph (DAG), $G(V, E)$ with a given scenario ($s$), where $V$ is a set of essential events $\{v_1, ... v_i, ... v_{|V|}\}$ and $E$ is a set of temporal ordering constraints between events $\{e_{ij}\}$ which means that the events $v_i$ must precede the event $v_j$ ($v_i \prec v_j$).\footnote{Technically, \data is a transitive reduction of a DAG. In short, transitive reduction of $G$ does not have any short cut edges between nodes. In \task, we add a single \textit{root} node ($v_r$) and scenario ($s$) as a unique leaf node.}
DAGs effectively encode the partial-ordering of core events--crucial for representing events which can be performed in any order.
For example, in a \textit{bake a cake} scenario, one can ``gather the ingredients'' and ``turn on the oven'' in any order (Figure~\ref{fig:fig1}).
We emphasize that scripts should not include non-core events such as discourse related events (e.g., reporting verbs) as \newcite{chambers-2017-behind} proposed. 
In \task, we also exclude alternative events in a \data DAG. For example, in a \textit{bake a cake} scenario, ``get ingredients'' and ``buy ingredients'' are alternative events with each other because either one is only necessary in the scenario. 
By excluding alternative events, we can resolve ambiguity of the edges in partial order structure as temporal relations or alternative paths. \newcite{regneri-etal-2010-learning} and \newcite{modi-etal-2016-inscript} do not discriminate this ambiguity.\footnote{In \task, we focus on events and the (partial-) ordering, and we leave participants/arguments identification for future work.}

With the definition, we introduce \task task in two complementary settings: script edge prediction and entire script generation.

\paragraph{Edge Prediction} 
The script edge prediction task is to predict a set of partial-ordered edges ($E$) of the script $G(V, E)$, given a scenario and a set of unordered intermediate events $v \in V$.

\paragraph{Script Generation}
The script generation task is to predict a partial order script $G(V, E)$, but only the scenario is given. 
Models are additionally expected to \textit{generate} events ($V$) in natural language.

%% file: src/datasets.tex
\section{Datasets}
\label{sec:datasets}

\paragraph{Source of Scenarios}
We collected scenarios from ROCStories~\cite{mostafazadeh-etal-2016-corpus}, DeScript~\cite{wanzare-etal-2016-crowdsourced}, and VirtualHome~\cite{puig2018virtualhome}. 
As ROCStories consists of sentences instead of scenarios, we extract phrases that match the manually curated patterns \textit{``want(ed) to ...",  ``need(ed) to ...", ``look(ing) to ..."} and that do not include personal pronouns or person's name. 
The 2,565 scenarios we collected include both high-level long-term ones (e.g., open a small business) and fine-grained short-term ones (e.g., sign into an email account). 
DeScript consists of 40 daily scenarios (e.g., making coffee) and we use all of them.
VirtualHome is constructed to learn activities interactively in a household in a 3D simulated world. 
It has 233 indoor tasks (e.g., turn on light) and we include them as scenarios.

\paragraph{Crowdsourcing \task}
For the collected scenarios, we crowdsource the corresponding \task on the Amazon Mechanical Turk. 
Our crowdsourcing procedure is similar but simplified method to \cite{Ciosici2021MachineAssistedSC}.
First, crowdworkers are required to describe five to seven \textit{core events} that they are essential for the given scenario~\cite{chambers-2017-behind} with the estimated time it takes to complete each event. 
In the second question, they are asked to sort them in possibly partial order (=DAG), which represents the \data for the scenario.

Due to the complex nature of this crowdsourcing procedure, it is crucial to maintain the quality. 
To identify and filter out noisy instances, two different workers are asked to sort the same set of events in partial order (i.e., the same as the second question described above). 
According to our manual analysis, we decided to retain scripts that have at least 65.0 F1 score between the workers.\footnote{In our crowdsourcing tasks, we maintained a pay rate of 12\$/hr or higher. For example, crowd workers were paid \$0.8 for the script creation and \$0.4 for the validation.}
To collect \task with both micro and macroscopic scenarios, we iteratively picked two adjacent events in the DAGs and use them as a source of finer-grained scenarios.

\begin{figure}[t]
\centering
\centerline{\includegraphics[width=1.0\columnwidth]{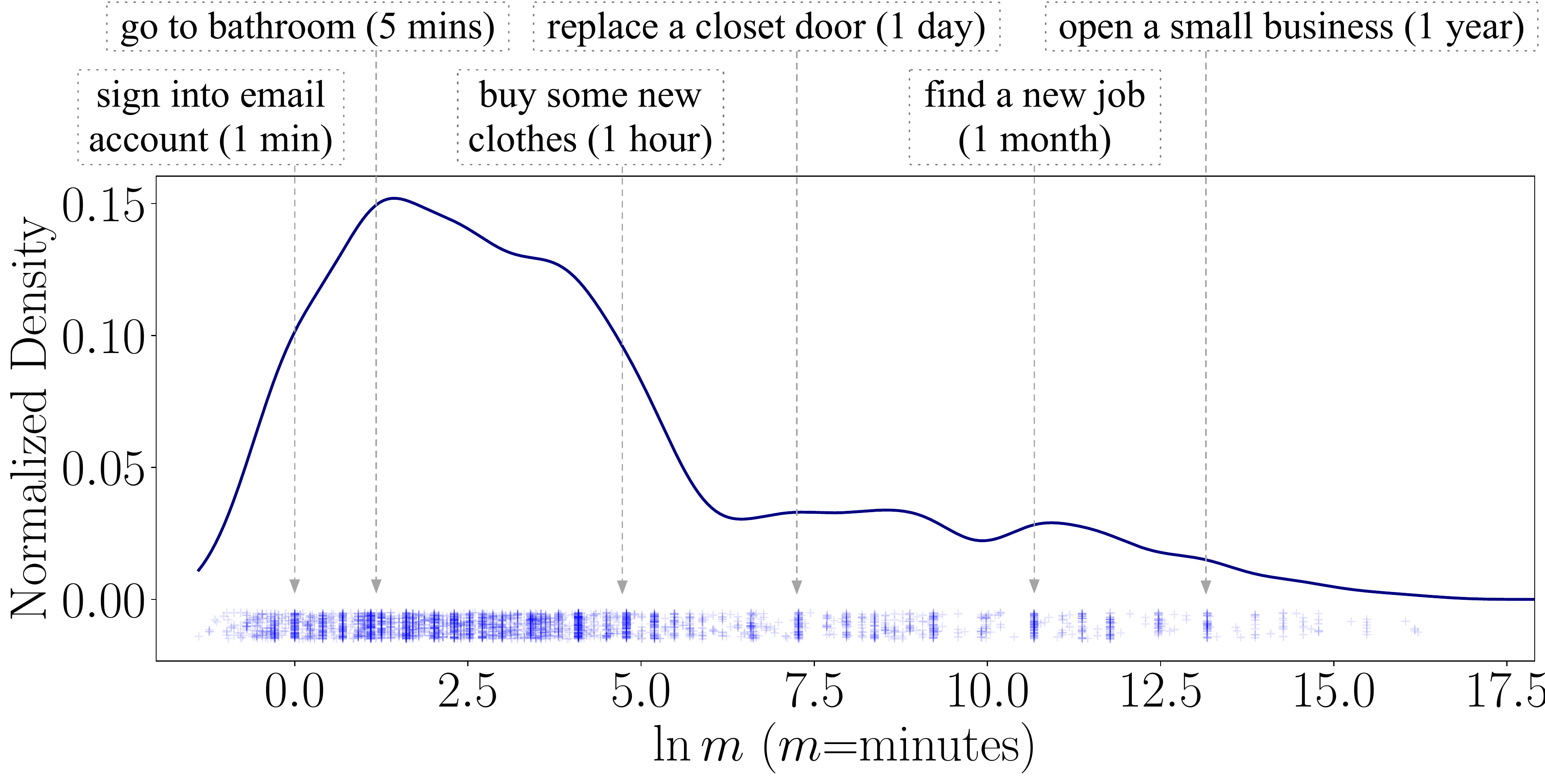}}
\caption{Normalized histogram of time duration in \task dataset. We see the dataset contains scripts with various time granularity.}
\label{fig:duration-kde}
\end{figure}

\paragraph{Dataset Statistics}
In total, we collected 6,414 valid scripts that include 311,502 pairs of events, and we split the \data into training (3,252 scenarios), development (1,085), and test set (2,077).
The training and development sets consist of scenarios collected from ROCStories, and the test set consists of those from ROCStories, DeScript, and VirtualHome. This helps us evaluate in- and out-of-domain performance.

The average number of events in \data scenarios is 5.45 and the maximum degrees of DAGs in the training set are distributed as follows: 
2,198 scripts (67.6\%) for degree 1, 
915 scripts (28.1\%) for degree 2, 
108 scripts (3.3\%) for degree 3, 
31 scripts (0.9\%) for degree 4 and above.

Figure~\ref{fig:duration-kde} shows the normalized histogram of the typical time to take for each script in \task dataset. 
Most of the scripts take between a minute and an hour (e.g., ``go to bathroom'', ``buy some new clothes''), while there are a reasonable amount of high-level long-term scripts (e.g., ``find a new job'', ``open a small business'').

%% file: src/task-edge.tex
\section{\task Edge Prediction}
\label{sec:edge-prediction}
\subsection{Models}
\label{sec:models-edge}
For the \task edge prediction task (\S\ref{sec:definition}), we implement a two-step approach baseline (\textit{pairwise model}) and compare it with our proposed end-to-end neural method (\modeledge).

\paragraph{Pairwise Model}

We implement a two-step baseline where we train a binary classifier to predict the precedence between pairs of events, followed by building a partial order script $\hat{G}$ by aggregating the predicted relations across all pairs of events. 

Formally, the classifier takes a pair of events ($v_i$, $v_j$) and predicts the precedence ${\hat{e}_{ij}}$ -- i.e. the event $v_i$ precedes ($\prec$) $v_j$ .
\begin{align}
\hat{e}_{ij} = p(v_i \prec v_j \vert v_i, v_j)
\end{align}

Scores by the classifier are used as weights to create an adjacency matrix of $G$ which is then automatically converted into a partial-order script with heuristics -- when $G$ contains a cycle, we iteratively remove edges by choosing the one with minimum weight until we get a valid DAG.

\paragraph{\modeledge}
\begin{figure}[t]
\centering
\includegraphics[width=0.9\columnwidth]{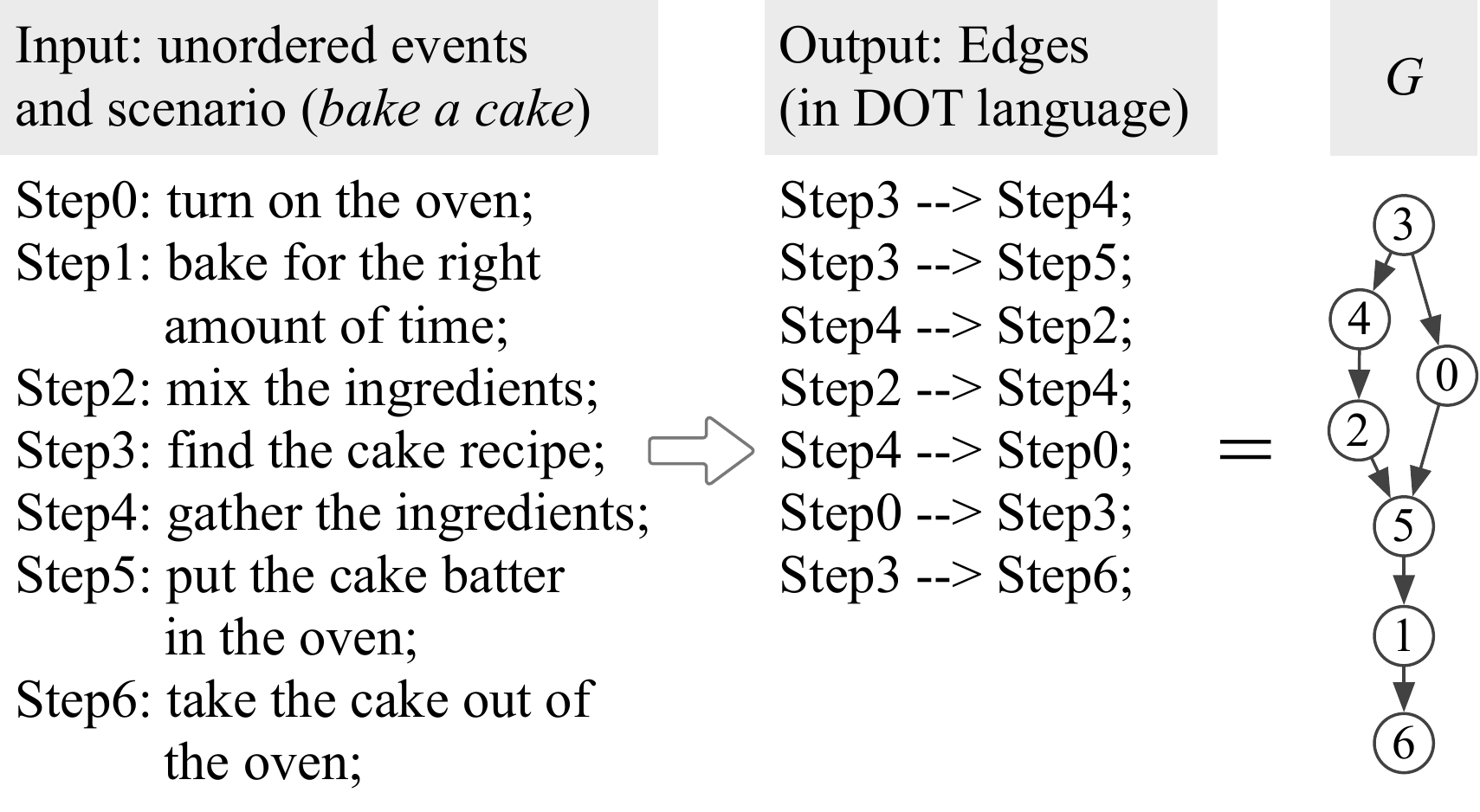}
\caption{Example of input and output for the \modeledge model. The input is a flattened sequence of events, and the output is a flattened sequence of edges of the (predicted) partial-order script.}
\label{fig:proscript-edge}
\end{figure}

We propose an an end-to-end neural model, which takes all the (unordered) events ($v$) and the scenario ($s$) as the input ($x$) and predicts the edges ($\hat{E}$) in a partial-order script ($\hat{G}$) at one time. 
To represent $\hat{E}$ in a linear format ($y$), we use DOT, a graph description language as shown in Figure~\ref{fig:proscript-edge}.\footnote{\newcite{Madaan2020NeuralLM} have previously shown that finetuned LMs can generate valid DOT language.}
By flattening the nodes and edges of $G$ (and $\hat{G}$), we apply neural encoder-decoder models.
Formally, flattened unordered events and scenario as $x$ are embedded as continuous representation (emb($x$)) by the encoder, then the decoder will generate tokens ($y$) as follows: 
\begin{align}
\label{eq:encdec}
p(y_1, \dots, y_N \vert x_1, \dots, x_M) &= \\
 \prod_{n=1}^{N}p(y_n \vert \text{emb}(x_1,\dots,x_M)&, y_1,\dots, y_{n-1}). \nonumber
\end{align}

Compared to the pairwise model, the \modeledge model uses information from all the events jointly to build partial-order script with a broader context.

\begin{table*}[t]
\small
\begin{tabular}{l|ccc|ccc|ccc|ccc}
\hline
            & \multicolumn{3}{c|}{dev} & \multicolumn{3}{c|}{test (all)} & \multicolumn{3}{c|}{test (in domain)} & \multicolumn{3}{c}{test (out domain)} \\ \hline
Models      & F1     & P      & R      & F1     & P       & R        & F1       & P        & R        & F1       & P        & R        \\ \hline
random      & 21.30 & 21.08 & 21.72 & 21.03 & 21.00 & 21.26 & 20.57 & 20.52   & 20.84    & 21.32    & 21.27    & 21.58 \\ \hdashline 
Pairwise (RoB) & 65.75 & 67.05 & 64.71 & 61.29 & 62.85 & 60.06 & 63.25 & 64.97 & 61.89 & 59.06 & 60.44 & 57.98 \\
Pairwise (T5)  & 70.96 & 71.93 & 69.76 & 67.64 & 69.44 & 66.18 & 69.50 & 71.41 & 67.96 & 65.51 & 67.20 & 64.16 \\ \hdashline
\texttt{proScr}(11B-100)  & 56.05 & 56.58 & 55.75 & 52.26 & 52.91 & 51.89 & 54.98 & 55.67 & 54.59 & 49.16 & 49.76 & 48.83 \\ 
\texttt{proScr}(11B-1k)   & 65.98 & 66.49 & 65.71 & 60.55 & 61.24 & 60.15 & 64.64 & 65.40 & 64.20 & 55.89 & 56.51 & 55.54 \\ 
\texttt{proScr}(Large)    & 66.25 & 66.89 & 65.83 & 63.64 & 64.22 & 63.27 & 65.76 & 66.38 & 65.35 & 61.23 & 61.76 & 60.91 \\
\texttt{proScr}(11B-all)  & \bf{78.20} & 78.48 & 78.14 & \bf{75.71} & 75.93 & 75.72 & \bf{77.75} & 78.03 & 77.71 & \bf{73.37} & 73.54 & 73.46 \\ \hline
Human            & 89.32 & 89.60 & 89.21 & 89.28 & 89.91 & 88.86 & 90.04 & 90.54 & 89.74 & 88.71 & 89.44 & 88.18 \\ \hline
\end{tabular}
\caption{Results for \task edge prediction task. In this table, \task refers to \modeledge.}
\label{tab:result-edge}
\end{table*}

\subsection{Evaluation Metrics}
Given $G(V, E)$ as a predicted (partial order) script and $\hat{G}(V, \hat{E})$ as the correct (oracle) script, the F1 score is defined as follows:
\begin{align*}
    \text{Precision} &= \frac{|E\&\hat{E}|}{|\hat{E}|}, \ 
    \text{Recall}    =  \frac{|E\&\hat{E}|}{|E|}          \\
    \text{F1 score}  &= \frac{2*\text{Precision}*\text{Recall}}{\text{Precision} + \text{Recall}}.
\end{align*}

\subsection{Experiments}
\label{sec:experiments-edge}

\paragraph{Setup}
For the binary classifier (pairwise model), we use two variants of the Transformer~\cite{vaswani2017attention}: RoBERTa-large~\cite{liu2019roberta} and T5-11B~\cite{2020t5}.

When training (i.e., fine-tuning) RoBERTa,\footnote{We used the implementation from Huggingface Transformers~\cite{Wolf2019HuggingFacesTS}.} we use a grid-search for choosing the best hyper-parameters from the best performed model on the development set: epochs \{1, 2, 3\}, learning rate \{1e-5, 1e-6, 1e-7\}, batch size \{16, 24, 32\}.
For training the T5 model as the pairwise model, we followed a default set of hyper-parameters that are recommended in~\newcite{2020t5}.

For the \modeledge model, we use the T5 with different model sizes (Large and 11B) and training sizes (100, 1k, and all 3.2k)  to see how these factors affect the performance.\footnote{We also used BART~\cite{lewis-etal-2020-bart}, but we found that BART did not perform well on this task.} 
We followed a default set of hyper-parameters for the T5 models.

\begin{figure}[t]
\centering
\centerline{\includegraphics[width=0.5\textwidth]{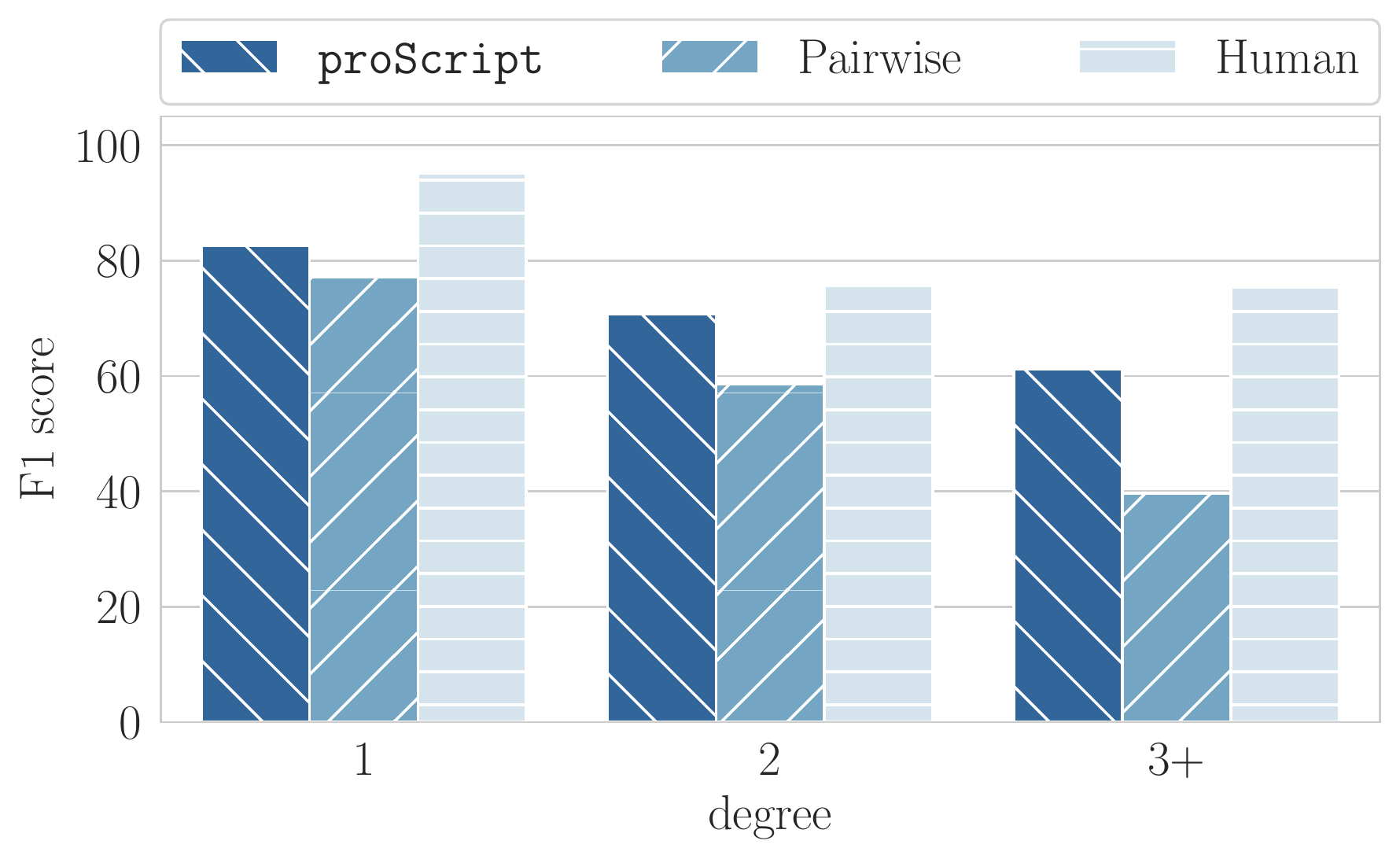}}
\caption{Model performance by pairwise (T5-11B), \modeledge (T5-11B) and human according to the maximum degree of the script (DAG).}
\label{fig:degree}
\end{figure}

\paragraph{Results}
The results are shown in Table~\ref{tab:result-edge}.
We find that the pairwise and \modeledge models significantly outperform the random baseline where the edges are randomly assigned.
The \modeledge T5-11B model outperforms the pairwise T5-11B model.
This indicates that the \modeledge model benefits from a larger context from the input to predict edges more accurately, although there is still a significant room for improvement toward human-level performance.\footnote{We find that 99\% of the outputs from \modeledge are valid DOT language.}
Regarding the difference between in and out of domain, we find that the in-domain performance is higher than the out-of-domain performance, whereas human performance is robust regardless of the domain difference.
We also see that the training set (100, 1k, all) and model sizes (Large, 11B) significantly affect the performance of \modeledge.

Figure~\ref{fig:degree} shows the performance of the pairwise (T5-11B) model, \modeledge (T5-11B) and human according to the (maximum) degree of the script DAGs.
We find that scripts with higher degree are more difficult to predict for both \modeledge and pairwise models, whereas human shows smaller decrease for predicting higher-degree scripts.

%% file: src/task-gen.tex
\section{\task Generation}
\subsection{Models}
\label{sec:models-gen}

\begin{figure}[t]
\centering
\includegraphics[width=1.0\columnwidth]{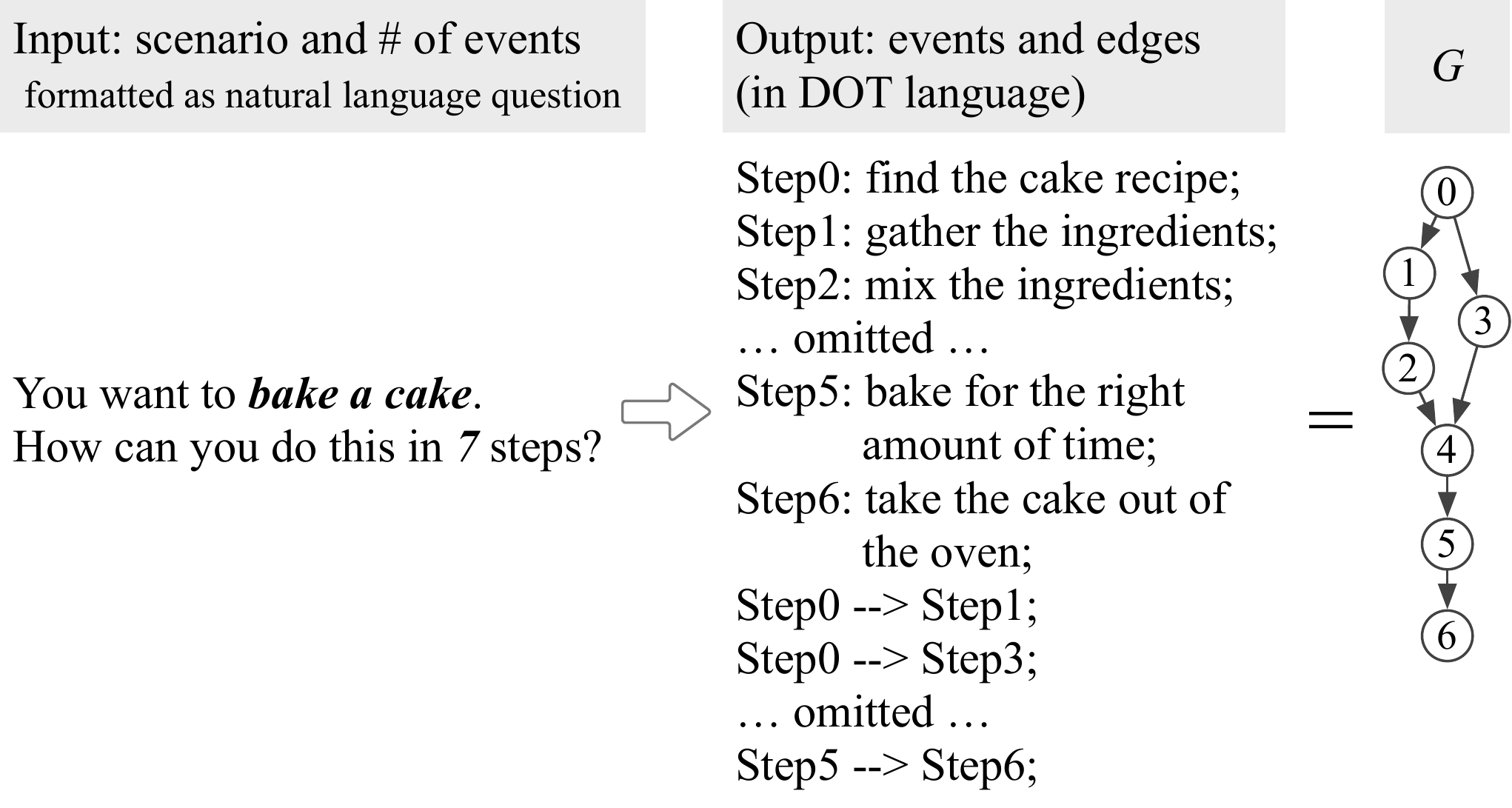}
\caption{Example of input and output for \modelgen. The input is a scenario and number of events to generate in natural text format, and the output is a sequence of events and edges of the script.}
\label{fig:proscript-gen}
\end{figure}

\paragraph{\modelgen}
The \task generation task combines natural language generation (i.e. generating events in natural language) with structure prediction over the generated events (i.e. organizing the events into a DAG).
Our approach (\modelgen) is to formulate it as an end-to-end problem, similar to the \modeledge for the \task edge prediction task (\S\ref{sec:models-edge}).
Given a scenario ($s$) and the number of events to generate in the script, \modelgen generates events and edges for the partial-order script ($\hat{G}$) in DOT language (Figure~\ref{fig:proscript-gen}).
Formally, we use the same encoder-decoder framework (eq.\ref{eq:encdec}) except that a scenario and number of steps to generate are described in natural text as $x$ and the decoder is expected to \textit{generate} events as well as the edges (as $y$) in the script.

\paragraph{Transfer learning from WikiHow data}
Transfer learning often helps improve the performance when it is (pre-)trained on a similar task~\cite{peters-etal-2018-deep,devlin-etal-2019-bert}.
As additional resource for pre-training \modelgen, we use procedural texts extracted from WikiHow,\footnote{\url{https://www.wikihow.com/}} which contains 130k instances of a sequence of essential steps for a given topic in various categories (e.g., health, finance, hobbies, etc.).
It is important to note that all the procedures in WikiHow are formatted as \textit{sequences} rather than a partial-order. We refer to this approach as \modelgentransfer.

\paragraph{Pipeline approach}
An alternative approach is to use \modelgen followed by the \modeledge model. The approach relies on \modelgen to generate a set of events but allows to fix the predicted edges via the \modeledge model.
We refer to this approach as \modelgenpipe, and study whether it can improve the performance over \modelgen.

\subsection{Evaluation Metrics}
\newcite{chambers-2017-behind} emphasizes the importance of human annotation for evaluating script knowledge.
However, human evaluation for the \task generation task is challenging because it involves natural text generation and structured prediction.
As in the text generation tasks such as machine translation and text summarization, there are several possible correct answers. 
Therefore, we use two complementary evaluation metrics for the \task generation task: (i) graph edit distance, and (ii) pairwise comparison. These are the absolute and relative measures of performance, respectively.
Graph edit distance~\cite{abuaisheh:hal-01168816} computes the distance between two graphs.
Formally, given two graphs $G_1$ and $G_2$,
\begin{align*}
    \text{GED}(G_1, G_2) =\min_{G_1\xrightarrow[]{d_1, ..., d_k}G_2} \sum_{i=1}^{k} \text{cost}(d_i)
\end{align*}
where $d_1,\ldots,d_k$ is a list of graph edit operations from $G_1$ to $G_2$.
The operations include deletion, insertion, and replacement for vertex and edge.
Each operation has its cost and we set the cost to be 1 for all the operations in our evaluation for simplicity.
We use the graph edit distance between a model-generated script and the revised script by human annotators.
The graph edit distance is indicative of the quality of the generated scripts; higher-quality scripts must have smaller graph edit distances to the gold-standard (i.e. they require a smaller number of human revisions).
In addition, we also employ pairwise human judgments where we ask human annotators to compare the scripts generated by \modelgen with those from the other approaches. 
\begin{table*}[ht]
\small
\center
\begin{tabular}{llccccccc}
\hline
Split & \multicolumn{1}{l|}{Models}        & \multicolumn{1}{c|}{Edit Dist} & V-Del & V-Ins & V-Rep & E-Del & E-Ins & E-Rep \\ \hline
    & \multicolumn{1}{l|}{\modelgen}         & \multicolumn{1}{c|}{\textbf{4.73}} & 0.426 & 0.192 & 0.581 & 1.558 & 1.308 & 0.671 \\
dev & \multicolumn{1}{l|}{\modelgentransfer} & \multicolumn{1}{c|}{4.79}      & 0.337 & 0.195 & 0.679 & 1.491 & 1.281 & 0.775 \\
    & \multicolumn{1}{l|}{\modelgenpipe}     & \multicolumn{1}{c|}{4.88}      & 0.397 & 0.159 & 0.560 & 1.705 & 1.407 & 0.661 \\ \hdashline
    & \multicolumn{1}{l|}{Human}            & \multicolumn{1}{c|}{2.78}      & 0.155 & 0.161 & 0.144 & 1.123 & 1.011 & 0.199 \\ \hline
    \\ \hline
      & \multicolumn{1}{l|}{\modelgen}         & \multicolumn{1}{c|}{\textbf{4.97}} & 0.581 & 0.142 & 0.656 & 1.668 & 1.184 & 0.709 \\
  test & \multicolumn{1}{l|}{\modelgentransfer}        & \multicolumn{1}{c|}{5.38}      & 0.438 & 0.213 & 0.775 & 1.713 & 1.402 & 0.835 \\
    & \multicolumn{1}{l|}{\modelgenpipe}         & \multicolumn{1}{c|}{5.41}      & 0.594 & 0.143 & 0.671 & 1.880 & 1.292 & 0.787 \\ \hdashline
    & \multicolumn{1}{l|}{Human}            & \multicolumn{1}{c|}{2.98}      & 0.168 & 0.149 & 0.130 & 1.276 & 1.074 & 0.189 \\ \hline
    \\ \hline
      & \multicolumn{1}{l|}{\modelgen}        & \multicolumn{1}{c|}{\textbf{4.57}} & 0.513 & 0.158 & 0.633 & 1.471 & 1.108 & 0.687 \\
  test (in domain) & \multicolumn{1}{l|}{\modelgentransfer} & \multicolumn{1}{c|}{5.03} & 0.339 & 0.299 & 0.649 & 1.575 & 1.496 & 0.677 \\
    & \multicolumn{1}{l|}{\modelgenpipe}         & \multicolumn{1}{c|}{5.10} & 0.561 & 0.147 & 0.630 & 1.765 & 1.217 & 0.744 \\ \hdashline
    & \multicolumn{1}{l|}{Human}            & \multicolumn{1}{c|}{3.03} & 0.168 & 0.211 & 0.154 & 1.223 & 1.091 & 0.206 \\ \hline
     \\ \hline
      & \multicolumn{1}{l|}{\modelgen}         & \multicolumn{1}{c|}{\textbf{5.43}} & 0.659 & 0.124 & 0.681 & 1.894 & 1.270 &	0.735 \\
  test (out domain) & \multicolumn{1}{l|}{\modelgentransfer} & \multicolumn{1}{c|}{5.76} & 0.549 & 0.115 & 0.916 & 1.867 & 1.296 & 1.013 \\
    & \multicolumn{1}{l|}{\modelgenpipe}         & \multicolumn{1}{c|}{5.81}  & 0.659 & 0.116 & 0.795 & 1.961 & 1.267 & 0.941 \\ \hdashline
    & \multicolumn{1}{l|}{Human}            & \multicolumn{1}{c|}{2.91}  & 0.170 & 0.074 & 0.102 & 1.340 & 1.054 & 0.170 \\ \hline
\end{tabular}
\caption{Results for \task generation task (dev, test, in-domain test and out-of-domain test set). We measure the average graph edit distance between generated script and the two human revisions (lower the better). We also show the average number of each graph edit operation (\{Delete, Insert, Replace\} $\times$ \{Vertex, Edge\}). Random (edge) baseline shows 11.06 edit distance for the dev set and 10.95 for the test set.}
\label{tab:result-plan}
\end{table*}

\subsection{Experiments}
\label{sec:experiments-gen}

\paragraph{Setup}
For our \modelgen, we use T5-11B. 
Similarly to the \modeledge, we follow the default set of hyper-parameters recommended in~\cite{2020t5}.
For \modelgentransfer, we pre-train the \modelgen with the 130k procedures, and finetune it on the \task dataset.
For the \modelgenpipe, we first obtain the actions generated by \modelgen (ignoring the edges), and use the set of events as input for \modeledge, which is trained (see \S\ref{sec:experiments-edge}) to predict the edges.

As defined in \S\ref{sec:definition}, we use graph edit distance and pairwise judgments to evaluate the quality of the generated scripts. 
For computing graph edit distances, we select 500 scripts (250 for dev and test sets) and ask crowdworkers to revise the generated scripts as necessary (e.g., add/delete/replace the events and the edges).
We use the revised scripts as gold-standard.
Each script is revised by two annotators, and we compute the average of the graph edit distances.

In pairwise judgments, we compare the scripts generated by \modelgen with those from the other approaches. 
We randomly select 150 pairs, and ask three crowdworkers to judge whether the script generated by \modelgen is \textit{better}, \textit{worse}, or \textit{equal to} the other (i.e. transfer, pipeline, or human).
We use majority vote to decide the final pairwise human judgment between the two scripts.

\begin{figure}[t]
\begin{center}
\includegraphics[width=1.0\columnwidth]{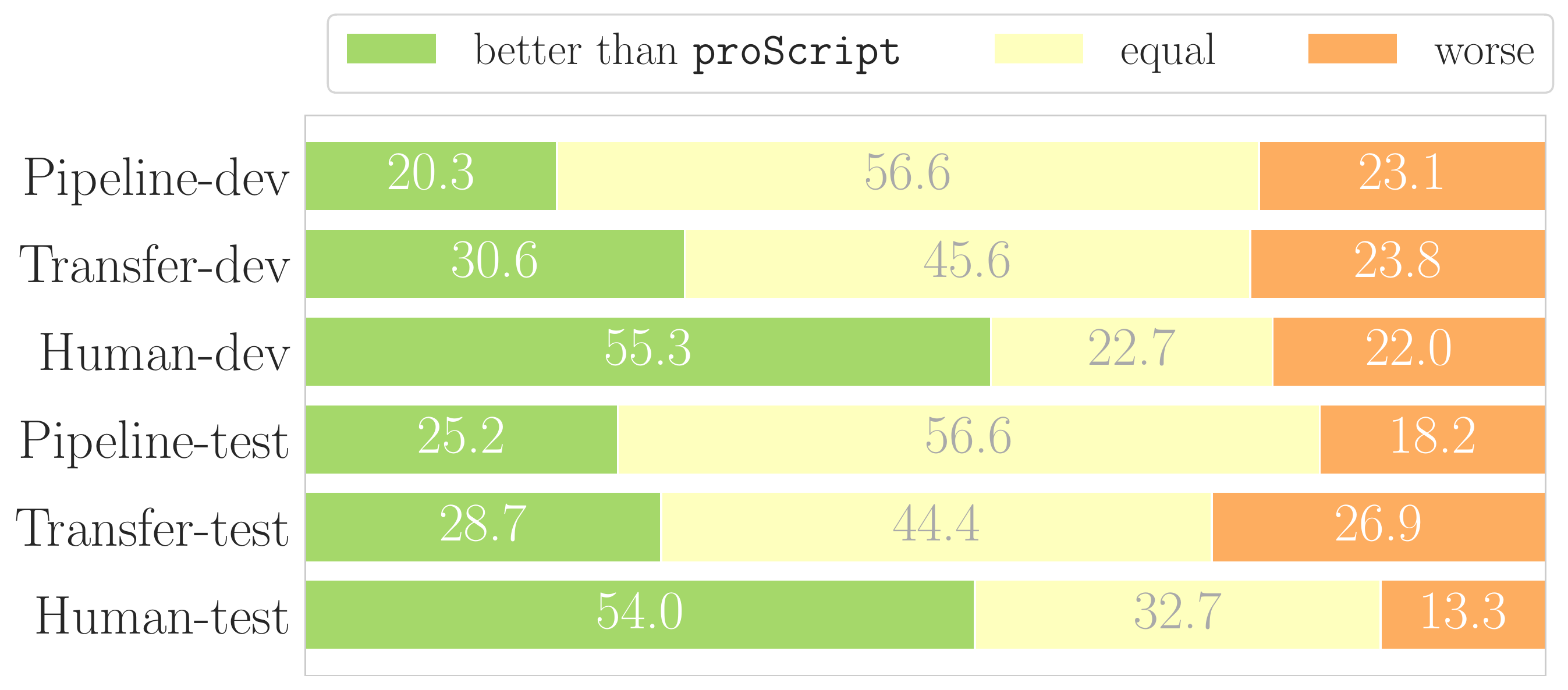}
\caption{Pairwise judgments (\%) between \modelgen and the other approaches.}
\label{fig:gen-pairwise}
\end{center}
\end{figure}

\begin{figure}[t]
\begin{center}
\includegraphics[width=1.0\columnwidth]{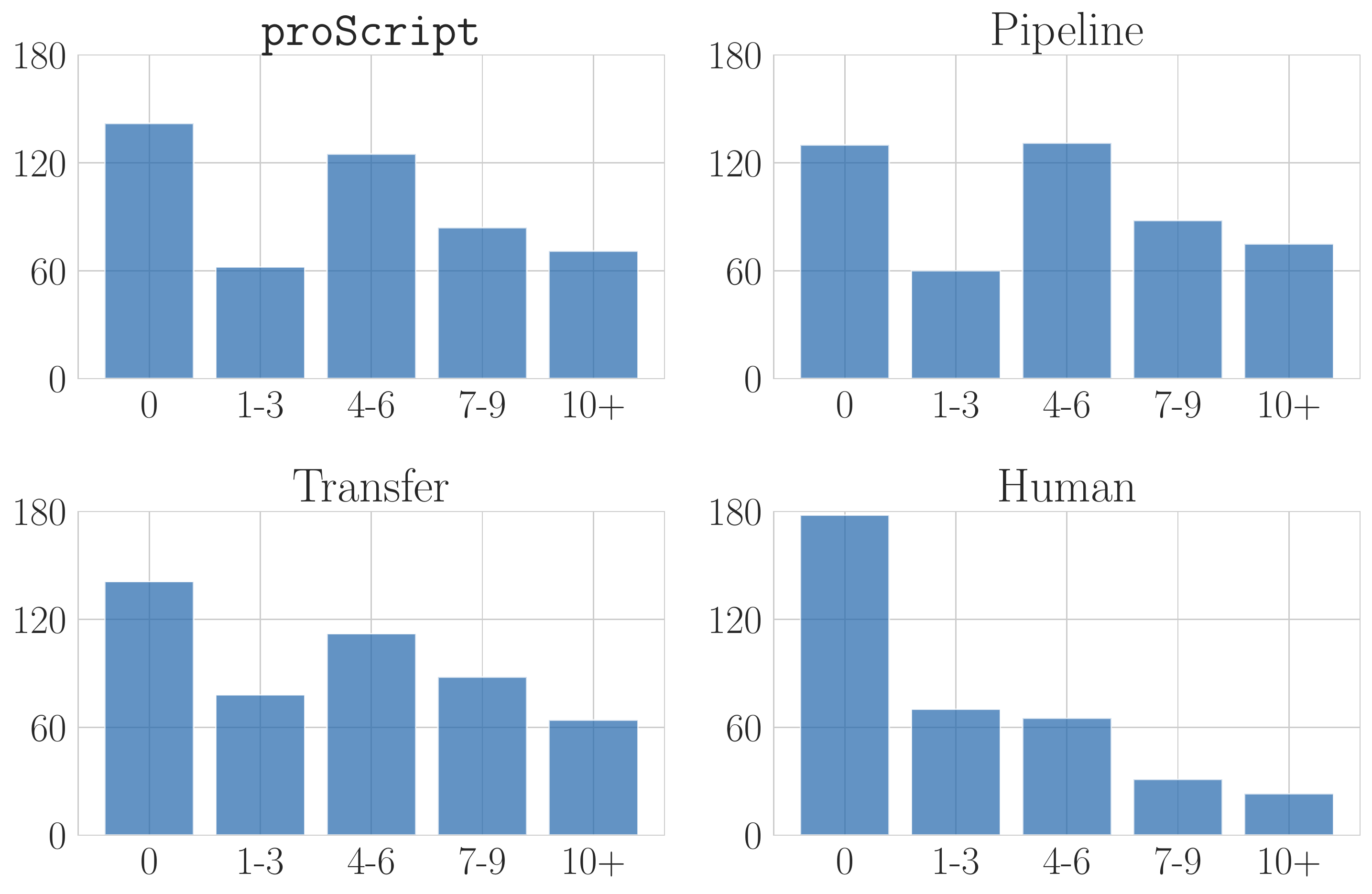}
\caption{Histograms of graph edit distance (in dev set). The number of scripts (y-axis) according to the (binned) graph edit distance (x-axis).}
\label{fig:gen-histogram}
\end{center}
\end{figure}

\paragraph{Results}
The pairwise judgment result is shown in Figure~\ref{fig:gen-pairwise}. 
We see that the pipeline and transfer models show slight preference over the \modelgen (except pipeline-dev), although the difference is not large. 
We also see that the transfer model constantly have more preference over the \modelgen than the pipeline model in both dev and test sets. 
Regarding the pairwise comparison with human-created plans, \modelgen still has a significant room for improvement toward human level. 

\begin{table*}[t]
\small
\begin{tabular}{p{0.28\columnwidth} | p{0.82\columnwidth} | p{0.82\columnwidth}}\hline
Revision types & generated script (subgraph) & revised script (subgraph) \\ \hline
missing event             & wait for the plane $\rightarrow$ exit the plane                                        & wait for the plane $\rightarrow$ get on the plane $\rightarrow$ exit the plane         \\ \hdashline
incorrect order            &  get off the car $\rightarrow$ drive to the zoo & drive to the zoo $\rightarrow$ get off the car                                          \\ \hdashline
irrelevant or redundant event & put clothes in dryer $\rightarrow$ place clothes into dryer $\rightarrow$ dry clothes & put clothes in dryer $\rightarrow$ dry clothes                                          \\ \hdashline
order by context & get a visa $\rightarrow$ ... $\rightarrow$ get off the plane $\rightarrow$ trip to a foreign country                                         & get off the plane $\rightarrow$  get a visa (on arrival) $\rightarrow$ trip to a foreign country      \\ \hdashline
granularity                & get out of the bed $\rightarrow$ go to the kitchen                                     & get out of the bed $\rightarrow$ open the bedroom door $\rightarrow$ go to the kitchen \\ \hdashline
paraphrased                & move into new apartment                                                                 & move to a new apartment \\ \hline
\end{tabular}
\caption{Examples for each revision type.}
\label{tab:plangen-error-analysis-examples}
\end{table*}

\begin{table*}[t]
\small
\center
\begin{tabular}{ll|r|r|r|r}
\hline
\multicolumn{2}{l}{Revision types}             & \multicolumn{1}{|c}{\task} & \multicolumn{1}{|c}{Transfer} & \multicolumn{1}{|c}{Pipeline} & \multicolumn{1}{|c}{Human} \\ \hline
\multicolumn{1}{l}{} & (edge) incorrect order             & 15.79                    & 21.62                         & 24.32                        & 10.00                     \\
\multicolumn{1}{l}{crucial errors} & (node) missing event & 5.26                     & 2.70                          & 2.70                         & 0.00                      \\
\multicolumn{1}{l}{} & (node) irrelevant/redundant event & 10.53                    & 13.51                         & 2.70                         & 0.00                      \\ \hdashline
\multicolumn{1}{l}{} & (edge) order by context & 31.58                    & 32.43                         & 40.54                        & 33.33                     \\
\multicolumn{1}{l}{minor revisions} & (node) granularity                 & 31.58                    & 24.32                         & 21.62                        & 26.67                     \\
\multicolumn{1}{l}{} & (node) paraphrased event & 0.00                     & 0.00                          & 5.41                         & 6.67                      \\ \hdashline
\multicolumn{1}{l}{wrong revisions}    & {}                 & 5.26                     & 5.41                          & 2.70                         & 23.33                     \\ \hline
\end{tabular}
\caption{Revision type distribution (\%) by each model. }
\label{tab:plangen-error-analysis}
\end{table*}

Table~\ref{tab:result-plan} shows the average graph edit distance between the generated script and the human revisions.
We find that neither transfer nor pipeline help to improve the graph edit distance over \modelgen, indicating that \modelgen is already a strong baseline (see examples in Appendix).
The reason of no improvement by the transfer approach may be because WikiHow consists of sequences rather than partially ordered steps.
No improvement by the pipeline approach indicates that the \modelgen can directly generate valid script in both events and edges.
Further studies for improvements are needed for future work.

In terms of the edit types, many of the edits are edge-related, suggesting that \modelgen and the variants are all good at generating events but struggles with ordering them.
Regarding in- and out-of domains in the test sets, we observe that \modelgen and the variants have slightly better performance for in-domain scripts than out-of domain, while human created scripts are not affected by domains.
These findings are consistent with the result in the edge prediction task (\S\ref{sec:experiments-edge}).

Figure~\ref{fig:gen-histogram} shows a histogram of the graph edit distance. It is evident that human created scripts are corrected less often than scripts generate by \modelgen, whereas the scripts from \modelgen and the variants often have a large number of edits (e.g., 4 or more).
It is interesting to see that fewer number of scripts have 1 to 3 edits (except scripts created by human).
The reason is because one simple revision tends to yield multiple graph edits (e.g., one node insertion yields multiple edge insertions).

\paragraph{Error Analysis} 
We performed manual error analysis for the scripts generated by each model.
We selected 40 random scripts that have non-zero graph edit distance and classified the human revisions into 7 types: (1) incorrect order, (2) missing event, (3) irrelevant/redundant event, (4) order by context, (5) granularity, (6) paraphrased event, and (7) wrong correction (examples are shown in Table~\ref{tab:plangen-error-analysis-examples}).
Approximately, the first three types indicate that the script has crucial errors, the next three types are trivial revisions where both generated and revised scripts are plausible.
The last type of revision is the one where the revised script is wrong (or worse). 

Table~\ref{tab:plangen-error-analysis} shows the statistics of each error type. We see that edge-related revisions are more frequent than node-related revisions. This is consistent with the results in graph edit distance. Overall, we find that minor revisions are more frequent than crucial errors, indicating that \modelgen and the variants generates reasonably good scripts. In contrast, crucial errors are quite rare in human created scripts, indicating a significant room for future innovation.

%% file: src/conclusions.tex
\section{Conclusions}
\label{sec:conclusions}
We show for the first time that pre-trained neural language models can be adapted to \textit{generate} partial order scripts.
We collect 6,400 partially ordered script from crowdsourcing (\data), which is substantially larger than prior manually crafted datasets.
With the \data dataset, we introduced two complementary task and models, providing the first demonstration that generative models can be successfully applied to script generation, although it is still below human performance.
We believe that \data dataset and models would advance future work on various NLP tasks such as story generation, machine comprehension, temporal reasoning, and high-level planning.

%% file: src/appendices.tex
\subsection{Plans generated by \modelgen}
\begin{figure*}[t]
\centering
\includegraphics[width=1.0\textwidth]{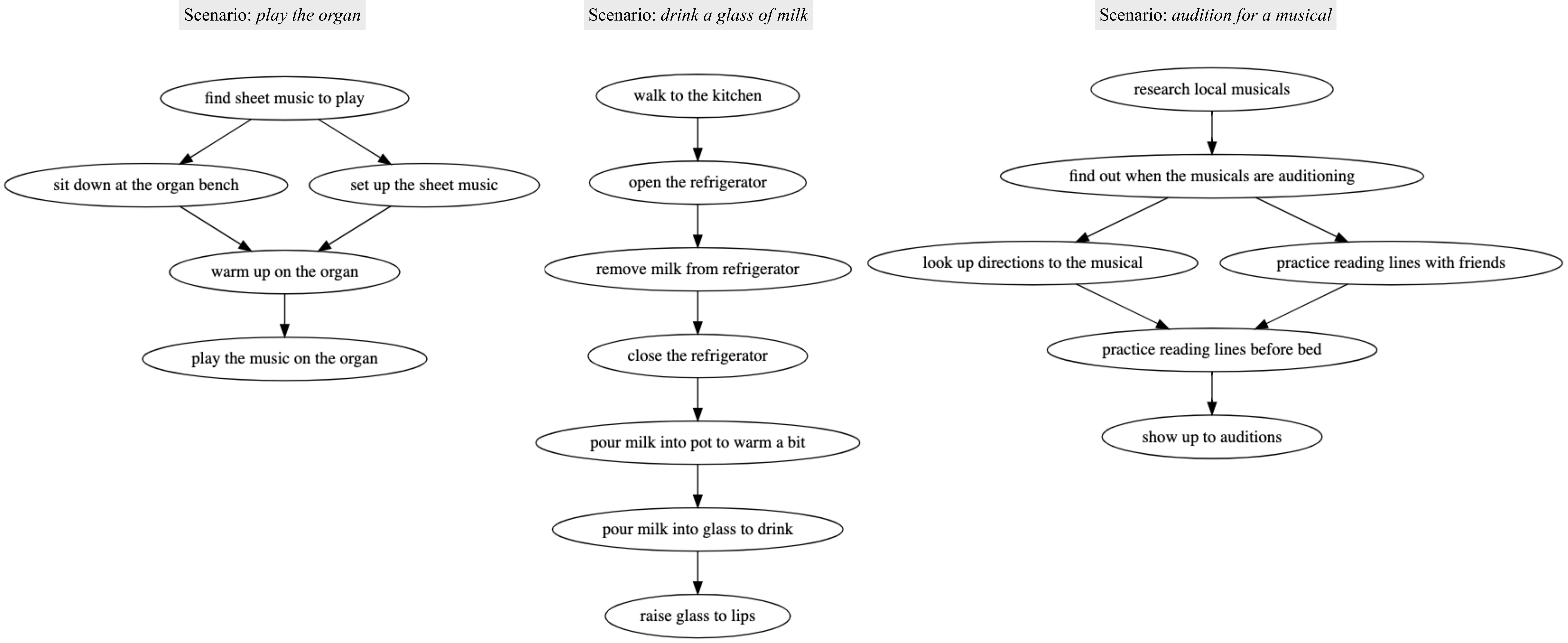}
\caption{Example scripts generated \modelgen.}
\label{fig:gen-examples}
\end{figure*}

We show some example scripts generated by \modelgen in Figure~\ref{fig:gen-examples}.

\subsection{Reproducibility}
For the POP model. we use a single CPU with 4GB memory. This model does not require any training procedure.
For training RoBERTa-large as a pairwise model, we use Quadro RTX 8000 (48GB memory), which takes around 4.5 hours to train a model.
RoBERTa-large consists of 355M parameters with 24 layers, 1,024 of hidden embedding size, and 16 of the attention heads.
T5-large model has 770M parameters with 24-layers, 1024-hidden-state, 4096 feed-forward hidden-state, and 16 attention heads.
T5-11B models  has 11B parameters with 24-layers, 1024-hidden-state, 65,536 feed-forward hidden-state, 128 attention heads.
We use TPU (v3-8) on google cloud platform. 
It takes 3 hours in average to train a edge prediction model, and 5 hours for plan generation models.